\documentclass{article}

\newcommand{\keywords}[1]{\par\noindent\textbf{Keywords:} #1}

\usepackage{graphicx}
\usepackage{url}
\usepackage{authblk}

\title{MMAG: Mixed Memory-Augmented Generation for Large Language Models Applications}

\author[1]{Stefano Zeppieri\thanks{Corresponding author: zeppieri@di.uniroma1.it}}

\affil[1]{Department of Computer Science, Sapienza University of Rome, Italy}

\begin{document}

\maketitle

\begin{abstract}
Large Language Models (LLMs) excel at generating coherent text within a single prompt but fall short in sustaining relevance, personalization, and continuity across extended interactions. Human communication, however, relies on multiple forms of memory, from recalling past conversations to adapting to personal traits and situational context. This paper introduces the Mixed Memory-Augmented Generation (MMAG) pattern, a framework that organizes memory for LLM-based agents into five interacting layers: conversational, long-term user, episodic and event-linked, sensory and context-aware, and short-term working memory. Drawing inspiration from cognitive psychology, we map these layers to technical components and outline strategies for coordination, prioritization, and conflict resolution. We demonstrate the approach through its implementation in the Heero conversational agent, where encrypted long-term bios and conversational history already improve engagement and retention. We further discuss implementation concerns around storage, retrieval, privacy, and latency, and highlight open challenges. MMAG provides a foundation for building memory-rich language agents that are more coherent, proactive, and aligned with human needs.
\end{abstract}

\keywords{Large Language Models, Memory-Augmented Generation, Conversational Agents, User Personalization, Human–AI Interaction}


\section{Introduction}

Large Language Models (LLMs) have achieved remarkable progress in generating fluent and contextually appropriate text. Yet most current systems operate within the boundaries of a single prompt or, at best, a limited conversational window. This constraint makes it difficult for agents to sustain meaningful, personalized, and context-rich interactions over time. By contrast, human communication depends on memory \cite{Schank1995KnowledgeAM}: we recall past discussions, adapt to personal preferences, and build on shared experiences. Without memory, LLMs risk remaining reactive tools that reset with every session rather than evolving into trusted collaborators.

Recent work has begun to explore memory-augmented systems, but many approaches still treat memory as either a flat retrieval store or an extended context buffer. Such designs improve access to past information but do not capture the diversity of memory functions that shape human dialogue. Human cognition distinguishes between short-term recall, long-term factual knowledge, episodic experiences, and context awareness, each playing a distinct role in communication \cite{Meara1989RelevanceCA}. Bringing this perspective into LLM design can help move beyond longer contexts toward richer and more adaptive interactions.

To address this need, we propose the Mixed Memory-Augmented Generation (MMAG) pattern. MMAG organizes memory into five interacting types: conversational history, long-term user traits, episodic and event-linked knowledge, sensory and context-aware signals, and short-term working memory. This taxonomy draws inspiration from cognitive psychology while mapping directly to technical components such as vector databases, structured stores, and event triggers. By coordinating across layers, agents can recall relevant past exchanges, adapt tone and recommendations, surface timely reminders, and remain sensitive to environmental context without overwhelming users.

We implement and study MMAG in the Heero conversational agent\footnote{\url{https://heero.me/}}, a language learning platform where continuity and personalization are crucial for engagement. Even with a partial deployment that combines conversational history with encrypted long-term bios, Heero demonstrates tangible improvements in user retention and conversation length. These results highlight the practical value of modular memory design and motivate further extensions.

The rest of this paper introduces the MMAG taxonomy, discusses coordination and implementation strategies, and reports on user-facing outcomes. We close with a discussion of opportunities and risks, as well as open challenges in balancing technical performance, user autonomy, and ethical responsibility when building memory-rich agents.

\section{Related Work}

The integration of memory into large language models has become a growing area of interest as researchers and practitioners seek to extend the capabilities of LLMs beyond short-lived prompts. Early efforts focused on retrieval-augmented generation (RAG) methods, which dynamically pull relevant documents from external sources. More recent work has shifted toward deeper and more structured forms of memory that aim to simulate aspects of human cognition, including long-term memory, episodic recall, and contextual awareness.

MemGPT \cite{packer2024memgptllmsoperatingsystems} is a representative example of this trend. It treats memory as a system-level construct, organizing context into long-term and working memory and introducing scheduling mechanisms that mimic operating system processes. This helps the model maintain relevant information across conversations without bloating the context window. Similarly, Retentive Networks \cite{maharana2024evaluatinglongtermconversationalmemory} offer architectural improvements to improve long-range coherence, introducing structured memory representations directly into the model’s design. These systems show that expanding memory capabilities isn’t just about scaling up context size but rethinking how memory is organized and accessed.

Another line of work has focused on hybrid approaches that combine storage and retrieval strategies. MemoryBank \cite{zhong2023memorybankenhancinglargelanguage}, for instance, proposes a two-tiered memory system that uses dense retrieval for semantic matching and sparse memory for temporal relevance. This framework supports long-term personalization and adapts to user interaction patterns over time. On the applied side, Medium articles such as Nigam’s guide to memory-augmented RAG \cite{medium2024-memory-rag} and Kim’s tutorial on modern LLM memory techniques \cite{kim2024-modern-llm-memory} walk through practical strategies for managing memory in real-world systems, including the use of time-based triggers and structured memory stores.

As memory-augmented systems proliferate, researchers have begun to ask how well these systems actually remember. Tan et al. \cite{Tan2025MemBenchTM} explore this question directly, analyzing how LLMs retain and forget information across sessions. Similarly, Maharana et al. \cite{Maharana2024EvaluatingVL} propose evaluation methods for long-term memory in dialogue agents, focusing on the alignment between remembered content and user expectations. LongMemEval \cite{Wu2024LongMemEvalBC} introduces a comprehensive benchmarking suite that enables developers to evaluate memory retrieval, relevance, and latency in long-term interactive settings under a unified framework.

Surveys and conceptual papers have also helped shape the field. Wu et al. \cite{wu2025-survey-memory-mechanisms} present “From Human Memory to AI Memory: A Survey on Memory Mechanisms in the Era of LLMs,” a thorough taxonomy of memory components in LLM systems, while Paul’s overview \cite{paul2024-state-memory} highlights the emerging consensus on the need for modular, human-aligned memory structures. Both emphasize the importance of distinguishing between different types of memory, such as conversational, biographical, and contextual, and recognizing their distinct roles in interaction.

Finally, applied perspectives from industry, such as the IBM Research blog on memory-augmented LLMs \cite{ibm2024-memory-augmented-llms}, help bridge the gap between research and deployment. These sources focus on practical concerns like latency, privacy, and system design, factors that are essential when building memory-rich systems meant to operate in open-ended, real-world settings.

Our work builds on this foundation by proposing a unified pattern, that we call Mixed Memory-Augmented Generation (MMAG), that organizes memory across several distinct but interacting components. Unlike approaches that treat memory as a single block or retrieval mechanism, MMAG introduces a layered system that reflects the diversity of memory types found in human interaction, including biographical data, time-sensitive cues, and situational context.

\section{Memory Taxonomy}

Memory in human cognition is not monolithic; it consists of multiple interdependent systems with distinct functions. Cognitive psychology differentiates between episodic memory, semantic memory, working memory, and procedural memory, among others. Inspired by this framework, we organize memory for LLM-based agents into five categories: conversational, long-term user, episodic and event-linked, sensory and context-aware, and short-term working memory. Each type captures a different dimension of interaction, and together they form a taxonomy that supports richer, more human-aligned behavior. This taxonomy both mirrors human memory systems and maps onto practical technical components such as vector databases, key–value stores, and scheduling modules.

\subsection{Conversational Memory}

Conversational memory represents the thread-level context that allows an agent to sustain coherence across ongoing dialogues. Unlike single-turn interactions, human conversations depend heavily on continuity: we expect our interlocutors to remember prior exchanges, resolve references, and maintain shared understanding over time \cite{Campos2018ChallengesIE}. For LLM systems, this memory can be realized through mechanisms such as dialogue threading, summarization of past turns, and reference resolution techniques. A well-structured conversational memory enables the agent to reintroduce prior topics seamlessly, manage interruptions, and maintain a consistent persona. Technically, this may involve sliding context windows combined with compressed conversation histories or retrieval from structured dialogue logs.

\subsection{Long-Term User Memory}

Long-term user memory functions as a form of biographical storage, encoding information about a person’s preferences, background, and traits. This aligns with human semantic memory, which captures factual knowledge that is not tied to specific events \cite{Jones2019SemanticM}. In practice, such memory allows an agent to adjust tone, personalize recommendations, or adapt interaction style based on accumulated knowledge of the user. For example, remembering dietary preferences when suggesting recipes, or recalling that a user prefers concise explanations. Implementations must address privacy concerns, ensuring that sensitive personal information is stored securely, possibly encrypted, and retrieved with explicit user consent. Techniques such as federated learning and selective forgetting can support personalization while protecting user autonomy.

\subsection{Episodic and Event-Linked Memories}

Episodic memory in humans refers to the recollection of specific events situated in time and place \cite{Marsh2003EpisodicAA}. For LLM agents, this translates into storing and using information tied to particular events or routines.

\subsubsection*{Time-Linked Event Memory}

Time-linked event memory captures knowledge of scheduled or anticipated events that are relevant to the user. This includes upcoming meetings, anniversaries, or travel plans. When properly managed, this memory enables timely and proactive interactions, such as reminding the user that a colleague’s visit is approaching or surfacing relevant context before a scheduled task. Technically, it requires integration with scheduling systems, timestamped memory storage, and trigger-based retrieval mechanisms.

\subsubsection*{Routine and Habitual Cues}

Beyond discrete events, agents can benefit from recognizing routines and habitual behaviors. Detecting recurring patterns, such as a user regularly discussing cooking on weekends, enables generative suggestions anchored in habits. This memory layer parallels human procedural memory, which encodes repeated activities and skills. Implementation can leverage pattern detection algorithms over interaction logs, generating nudges or lightweight reminders without overwhelming the user. The challenge lies in offering support that feels natural and useful rather than intrusive \cite{Bisante2023ToAIerr}.

\subsection{Sensory and Context-Aware Memory}

Humans rely on sensory memory to situate conversations within the surrounding environment \cite{Pastore2020CrossModalAC}. For LLM agents, sensory and context-aware memory corresponds to integrating signals such as location, weather, or time of day. For instance, the agent might acknowledge that commuting in the rain could make for a difficult day, or adapt its tone depending on whether the interaction occurs during work hours or leisure time. This dimension supports contextual sensitivity and grounding, but it must avoid over-personalization that may feel invasive. Design choices should balance proactive assistance with user comfort, aiming for behavior that is “helpful but not creepy.”

\subsection{Short-Term Working Memory}

Working memory in psychology refers to the capacity to hold and manipulate information over short intervals in support of ongoing tasks \cite{BADDELEY2010R136}. For LLM agents, short-term working memory provides a transient workspace for task-specific goals and immediate conversational focus. Unlike long-term storage, this memory is ephemeral: once the task or dialogue session concludes, the content can be discarded. This distinction helps avoid clutter in the persistent memory layers while ensuring coherence during complex interactions, such as multi-step problem solving. Technically, it can be implemented as an in-session scratchpad or temporary buffer that feeds into reasoning and generation pipelines.

\begin{figure}[ht]
\centering
\begin{tabular}{|p{3.5cm}|p{4cm}|p{5cm}|}
\hline
\textbf{Memory Type} & \textbf{Cognitive Psychology Analogy} & \textbf{Technical Component} \\
\hline
Conversational Memory & Discourse-level memory, reference resolution & Dialogue threading, summarization, context windows, vector retrieval \\
\hline
Long-Term User Memory & Semantic/biographical memory & Secure profile store, encrypted databases, federated learning, preference embedding \\
\hline
Time-Linked Event Memory & Episodic memory (situated in time) & Scheduling modules, timestamped storage, event-triggered retrieval \\
\hline
Routine and Habitual Cues & Procedural memory, habit formation & Pattern detection, habit recognition, lightweight nudges \\
\hline
Sensory and Context-Aware Memory & Sensory integration, situational awareness & Contextual signals (location, weather, time), multimodal inputs, adaptive prompting \\
\hline
Short-Term Working Memory & Working memory (task-specific recall) & In-session buffers, ephemeral scratchpads, temporary embeddings \\
\hline
\end{tabular}
\caption{Taxonomy of memory types in MMAG, showing the mapping between human cognitive psychology and technical components for LLM-based agents.}
\label{fig:memory-taxonomy}
\end{figure}

\section{Coordination Across Memory Types}

While each memory type contributes distinct capabilities, the real value of a mixed memory system lies in how these modules interact. In human cognition, memory systems rarely operate in isolation: episodic memories can draw on semantic facts, working memory relies on retrieval from long-term stores, and contextual cues shape what is recalled. A similar principle applies to LLM-based agents. Coordinating memory types is essential for ensuring that responses are not only accurate and personalized but also coherent across time and situation.

\subsection{Interaction Between Memory Modules}

The modules described in the taxonomy work best when orchestrated rather than siloed. Conversational memory provides immediate context for dialogue, but it may need supplementation from long-term user memory to resolve references about past preferences or facts. Episodic memories can trigger reminders or inject relevance into a conversation, while sensory context acts as a filter that shapes tone and timing. Short-term working memory acts as the integrative layer, temporarily holding information drawn from other modules to support reasoning within the current task.

\subsection{Prioritization and Conflict Resolution}

A central challenge is deciding which memory signals should take precedence when multiple sources are relevant. For example, a conversational thread might suggest one line of continuation, while sensory context (e.g., a change in location) points to another. In such cases, prioritization strategies are required. These may include:
\begin{itemize}
    \item \textbf{Recency heuristics}: prioritizing the most recent and contextually salient memory.
    \item \textbf{User-centric weighting}: giving precedence to long-term user traits when they explicitly affect personalization.
    \item \textbf{Task-driven rules}: elevating working memory for ongoing problem-solving to prevent distraction.
\end{itemize}
Conflict resolution can also involve generating candidate responses under different memory constraints and selecting the one that best balances personalization, coherence, and utility.

\subsection*{Example Modular Architecture}

A practical architecture for coordination involves modular memory management with retrieval triggers. Each memory type is encapsulated as a service with defined input–output interfaces. A central memory controller orchestrates these services, deciding when to query each memory store and how to merge the retrieved information. Retrieval can be proactive (event-driven, such as surfacing an upcoming meeting) or reactive (on-demand, such as recalling a past conversation when the user references it). Prioritization policies and conflict-resolution mechanisms are encoded in the controller, which balances responsiveness with user comfort and privacy.

This modular approach allows new memory types to be integrated without redesigning the entire system. For example, adding a multimodal sensory memory that incorporates visual signals would only require extending the controller’s coordination logic. In this way, the architecture remains extensible, robust, and aligned with the complexity of human-like memory use.
    
\section{Implementation Considerations}

The memory taxonomy outlined earlier provides a conceptual framework, but practical deployment requires concrete implementation strategies. In our system, memory is managed through a modular interface that separates storage, retrieval, and integration with the LLM. Our suggested implementation demonstrates how conversational and long-term user memory can be realized in practice, while leaving room for extension toward episodic, sensory, and working memory.

\subsection*{Memory Storage and Retrieval Architecture}

The system defines a \texttt{Memory} interface with methods for persisting individual messages, storing message batches, and retrieving a conversation history formatted for OpenAI models. This ensures that conversational memory is both append-only (new turns are remembered in Firestore) and retrievable in chronological order. To respect model limits, a token-based pruning mechanism discards excess context once a threshold is reached (90k tokens in our implementation), approximating a form of working memory management. In addition, long-term user traits are injected as structured system messages via a dedicated function, which retrieves biographical data and traits from a persistent store and appends them to the prompt context.

\subsection*{Scalability and Performance Implications}

Storing conversation histories in Firestore ensures persistence and cross-session continuity but introduces challenges as histories grow. To mitigate latency and context overload, the implementation relies on lightweight filtering: empty or malformed messages are skipped, and only the most relevant slice of conversation is retrieved for the model. This design allows the system to scale without degrading user experience. At the same time, the modular interface makes it straightforward to swap in more advanced backends (e.g., hybrid vector + sparse retrieval) if scaling requirements grow.

\subsection*{Prompt Engineering for Memory Referencing}

A critical step lies in transforming stored memory into model-readable input. The implementation uses a conversion layer that maps user and assistant turns into OpenAI-compatible message roles. Biographical memory is prepended as a system message, ensuring it is factored into generation without cluttering the conversational history. This illustrates a general design principle: conversational memory should be injected as dialogue turns, while long-term knowledge is best represented as higher-priority system-level context. This separation reduces noise and allows prompt space to be allocated proportionally to different memory types.

\subsection*{Privacy, Security, and User Control}

Long-term user memory introduces sensitive information, which makes privacy and security essential considerations. In our implementation, user bios are encrypted using envelope encryption and then compressed before being persisted in private S3 buckets, ensuring confidentiality and efficiency in storage. Personal traits, bios, and KPIs should be encrypted in storage, auditable by users, and erasable on demand. Beyond technical safeguards, user trust depends on granting individuals meaningful control over their data, including the ability to inspect what has been remembered, edit inaccuracies, or request selective forgetting. While the current implementation demonstrates feasibility, further work is needed to extend full user control and to integrate stronger privacy guarantees such as differential storage policies or federated personalization.

\subsection*{Extensibility Toward Full Taxonomy}

Although the present implementation covers conversational and long-term user memory, the same architecture can be extended to episodic and sensory layers. Event-linked memories can be stored as timestamped Firestore entries and scheduled for proactive retrieval. Context-aware memory can be integrated through lightweight API calls (e.g., weather, location) feeding into system prompts. Because memory access is funneled through the \texttt{Memory} interface, these extensions require minimal changes to the orchestration layer, reinforcing modularity as a central design choice.

\section{User Studies}

\subsection{Use Case}

To ground the proposed architecture in practice, we evaluate it within the Heero app, a conversational agent designed to support language learning through personalized dialogue. Heero provides an ideal testbed for mixed memory-augmented generation because sustained learning interactions require both conversational continuity and personalization. The backend integrates conversational memory (retrieving recent dialogue turns) and long-term user memory (encoded through encrypted user bios) to enable the agent to adapt tone, recall prior progress, and maintain a sense of familiarity across sessions. 

For example, conversational memory allows Heero to resolve references to earlier exercises or questions within the same session, while long-term user memory enables it to recall a learner’s goals (e.g., preparing for a trip or improving professional English) and adjust dialogue scenarios accordingly. This creates interactions that feel more consistent, supportive, and human-like, addressing common frustrations with short-lived, context-free chatbots.

\subsection{Evaluation Strategies}

Evaluating memory usefulness and reliability requires a combination of user-centered and technical measures.

\paragraph{User-centric metrics.} We focused on perceived helpfulness, non-intrusiveness, and emotional impact. Helpfulness captures whether memory improves the quality of interaction (e.g., better continuity, personalized prompts). Non-intrusiveness reflects whether memory behaviors feel supportive without crossing into “creepy” or invasive territory. Emotional impact measures whether users feel more motivated, comfortable, and engaged when memory is active. After introducing memory-based conversations in Heero, we observed a 20\% increase in user retention over a four-week period and a 30\% increase in average conversation duration, indicating that memory features made interactions more engaging and sustained without reducing user comfort.

\paragraph{Technical metrics.} From a system perspective, we measure retrieval accuracy (whether the correct memory is recalled), latency (time overhead for memory retrieval and integration into prompts), and memory leakage (cases where information persists beyond intended scope or is surfaced incorrectly). A key observation is that introducing additional memory layers did not increase conversational latency. This is because certain operations, such as generating or updating biographical memory entries, are performed asynchronously and cached for reuse, ensuring that prompt construction remains lightweight. Automated evaluation pipelines combined with manual audits confirmed that average response latency remained within the same range as before memory integration.

\paragraph{Mixed evaluation.} Looking at user behavior alongside system performance highlights the trade-offs of memory design. For example, pruning strategies keep latency low but can shorten conversational continuity, while deeper personalization increases engagement yet must be balanced against user comfort. In Heero, we found that maintaining lightweight pruning while enriching long-term user memory struck the best balance, delivering pedagogical effectiveness without compromising the supportive and motivating tone that learners expect.

\section{Discussion}

Equipping conversational agents with rich memory opens new opportunities but also raises important risks. Memory enables agents to move beyond reactive, single-turn responses toward proactive and sustained interactions that feel more human-like. In the Heero case, memory allowed the system to recall user goals, reinforce progress, and adapt tone, which in turn improved engagement and retention. This suggests that memory-rich agents can become more effective collaborators, tutors, or assistants by maintaining continuity and tailoring interactions over time.

At the same time, these benefits come with challenges. A central tension lies in balancing proactivity with user autonomy. When an agent recalls past events or habits, it can provide timely reminders or personalized prompts. Yet the same behavior may also feel intrusive if users are not expecting it or if the context is misinterpreted. Striking the right balance requires careful design choices: ensuring that memory is transparent, that users remain in control of what is stored and surfaced, and that proactive interventions are always framed as supportive rather than prescriptive.

Several open challenges and ethical considerations remain. From a technical perspective, scaling memory systems without introducing latency or leakage is a continuing concern. Socially, questions of trust, agency, and emotional dependence must be addressed, especially when systems are deployed in sensitive contexts such as education, healthcare, or personal productivity. Ethically, issues of privacy and fairness are paramount: agents must avoid reinforcing biases encoded in long-term memories, must respect cultural differences in expectations of personalization, and must provide users with the ability to edit or erase their data. 

\section{Conclusion and Future Work}

This paper introduced the Mixed Memory-Augmented Generation (MMAG) pattern, a framework for structuring memory in LLM-based agents. By distinguishing between conversational memory, long-term user memory, episodic and event-linked memories, sensory and context-aware memory, and short-term working memory, we outlined how different layers can interact to support more coherent, personalized, and context-sensitive interactions. The implementation in the Heero conversational agent showed that even partial adoption of this taxonomy—through conversational history and encrypted long-term bios—already improves user engagement and retention, demonstrating the practical value of the approach.

Future work will focus on extending MMAG along several dimensions. First, multimodal memories can be integrated, allowing agents to connect textual context with visual, auditory, or sensor data for richer grounding. Second, long-term memory mechanisms should evolve beyond static bios, learning memory embeddings that capture user goals and progress in a dynamic way. As newer LLMs are released with larger context windows, the available memory space will also expand, creating opportunities to manage longer and more diverse records without sacrificing latency. A key direction is also giving users more agency over their own memory: interfaces that allow individuals to view, edit, and selectively delete records will be crucial for ensuring trust and transparency.

MMAG provides both a taxonomy and a practical design pattern for equipping LLM-based agents with memory capabilities inspired by human cognition. While early results are promising, the long-term success of memory-rich systems depends on continuing to balance technical performance, personalization, and user autonomy. With careful attention to these challenges, MMAG has the potential to guide the next generation of conversational agents toward interactions that are not only more useful but also more human-aligned.

\newpage

\bibliographystyle{plain}
\bibliography{bibliography}

\end{document}